%% file: particle_transformer.tex
\documentclass[conference]{IEEEtran}
\IEEEoverridecommandlockouts
\usepackage{cite}
\usepackage{amsmath,amssymb,amsfonts}
\usepackage{graphicx}
\usepackage{textcomp}
\usepackage{xcolor}
\usepackage{verbatim}
\usepackage[ruled,vlined]{algorithm2e}
\usepackage{tikz}
\usepackage{url}
\usetikzlibrary{shapes.geometric, arrows.meta, positioning}
\usepackage{amsthm}

\newcounter{heuristicctr}
\renewcommand{\theheuristicctr}{\arabic{heuristicctr}}
\newsavebox{\heuristicbox}

\newsavebox{\mybox}

\def\BibTeX{{\rm B\kern-.05em{\sc i\kern-.025em b}\kern-.08em
    T\kern-.1667em\lower.7ex\hbox{E}\kern-.125emX}}
\begin{document}

\title{Reconfigurable Computing Challenge: Transformer for Jet Tagging on Versal AI Engines}

\author{ 
\IEEEauthorblockN{Gram Koski, Sean Lipps, Zhenghua Ma, G Abarajithan, Ryan Kastner}
\IEEEauthorblockA{\textit{Department of Computer Science and Engineering} \\
University of California San Diego \\
La Jolla, CA, USA \\
\{gkoski, slipps, zhm007, agnaneswaran, kastner\}@ucsd.edu}
}

\newcommand\gram[1]{\textcolor{green}{\textbf{#1} -Gram}}
\newcommand\zhenghua[1]{\textcolor{blue}{\textbf{#1} -Zhenghua}}
\newcommand\aba[1]{\textcolor{cyan}{\textbf{#1} -Aba}}
\newcommand\ryan[1]{\textcolor{red}{\textbf{#1} -Ryan}}

\maketitle
\input{short-version/abstract}
\input{short-version/intro}

\input{short-version/background}

\input{short-version/method1}
\input{short-version/method2}

\input{short-version/results}
\input{short-version/conclusion}
\bibliographystyle{IEEEtran}
\bibliography{references}

\end{document}

%% file: short-version/abstract.tex
\begin{abstract}
Transformer-based models achieve strong performance for jet tagging at the CERN LHC, but deploying them in low-latency, resource-constrained trigger systems is challenging. We present an initial implementation of a quantized, integer-only transformer for jet tagging on the AMD Versal AI Engine (AIE), mapping dense and multi-head attention (MHA) layers to AIE tiles. The main contribution is a reusable software framework that represents transformer layers as composable AIE building blocks and automatically generates the corresponding Vitis graph code from a high-level Python model description. This framework provides a foundation for future research and is released as open-source software at \url{https://github.com/KastnerRG/particle_transformer_aie}.

\end{abstract}

%% file: short-version/intro.tex
\section{Problem and Motivation}

At the CERN Large Hadron Collider (LHC), proton--proton collisions occur at a rate of 40\,MHz, producing an enormous stream of particle jets that must be filtered in real time by the Level-1 Trigger (L1T) system\cite{LHCTriggerCloserLook, CMSL1TRealTime}. 
Jet tagging is the task of classifying jets according to their originating physics process.
While accurate jet tagging is essential for selecting interesting events, latency and throughput must be kept within tight budgets to enable real-time processing \cite{CMSL1TRealTime}. 
Recent work has shown that transformer-based architectures, such as the Particle Transformer (ParT), can achieve high accuracy by operating directly on sets of per-particle features~\cite{qu2022parttransformer}. 
However, such transformer models were deployed on GPUs in an offline setting, where millisecond-level latencies and high power consumption are acceptable.

\begin{figure}[ht]
    \centering
    \includegraphics[width=1\linewidth]{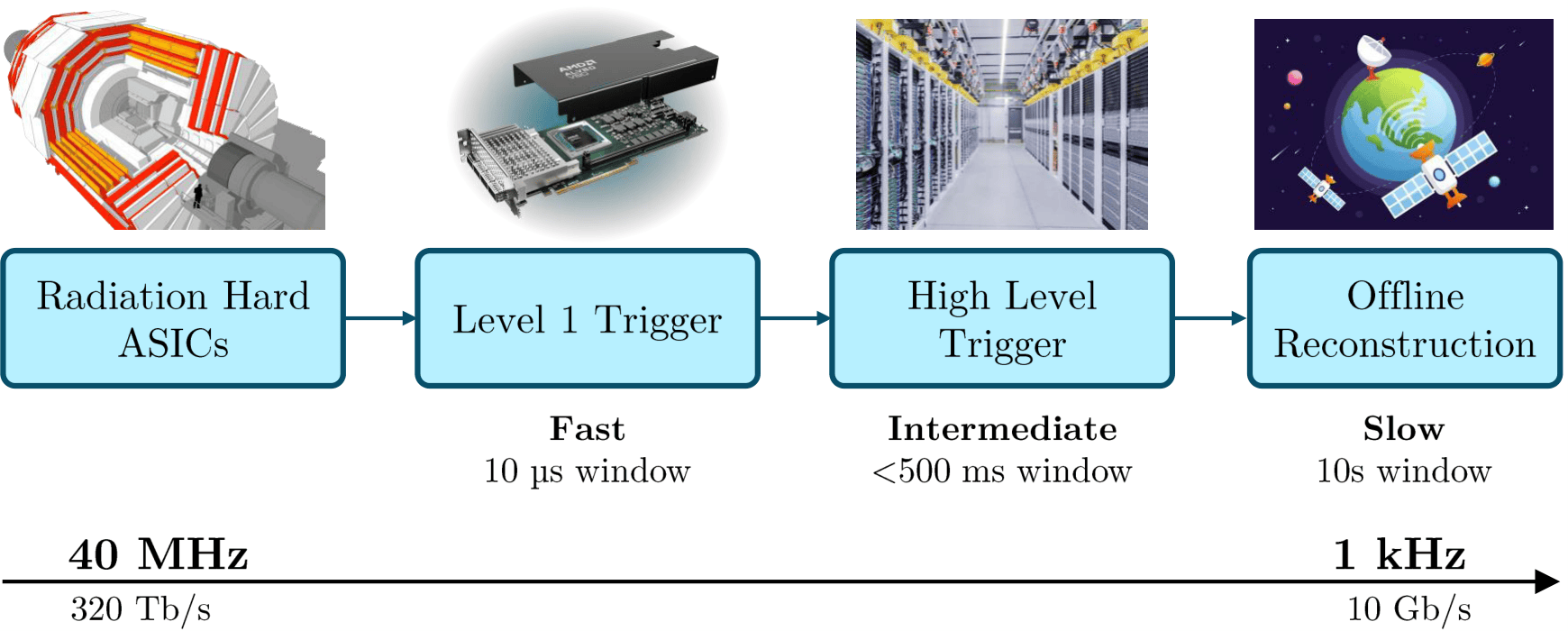}
    \caption{Trigger and inference latency scales at the LHC, from 40 MHz Level‑1 decisions to higher‑level triggers and offline analyses.}
    \label{fig:placeholder}
\end{figure}

For online triggering, the constraints are far more stringent: the L1 trigger must produce accept/reject decisions within a few microseconds while limiting the output rate to at most $\mathcal{O}(10^5)$ events per second and operating under tight on-detector power and resource budgets\cite{CMSL1TRealTime,CMSL1TDesign}. 
This makes deploying transformer models on traditional CPU/GPU platforms impractical for real-time edge inference. 
The AI Engine (AIE) array on AMD Versal SoCs offers a promising platform for low-latency, high-throughput ML inference.

In this work, we design and implement a transformer-based jet tagging accelerator
on the AMD Versal VCK190 platform. Our main contributions are as follows:
\begin{itemize}
  \item We introduce a reusable software framework that represents transformer
        layers as composable AIE building blocks and automatically generates
        the corresponding Vitis graph code from a high-level Python model
        description.
  \item We implement a fully quantized, integer-only transformer, including
        dense layers, residual connections, and an integer-only softmax,
        tailored to the arithmetic and memory constraints of the AIE array.
  \item We propose and evaluate a per-head multi-head attention (MHA) mapping that
        assigns heads and their sub-projections to parallel AIE tiles,
        achieving improved throughput
\end{itemize}

\begin{comment}
Rather than a fully optimized design, our
current implementation represents an initial, relatively straightforward mapping
of a quantized, integer-only transformer onto the AI Engine (AIE) array that
already achieves encouraging latency characteristics.
\aba{dont keep underselling your work. just mention your good points and leave the judgment at it} \zhenghua{If you really want to say it, put them as future work and why achieving them is practical and not hard. E.g., We present an initial, straightfoward mapping ...... achieves encouraging latency characteristics, and propose future work to ... optimizations. But don't be apologetic.}
\end{comment}

\begin{comment}
Putting the system diagram figure here so that it appears on page 2.
\end{comment}

\begin{figure*}[t]
  \centering
  \includegraphics[width=\textwidth]{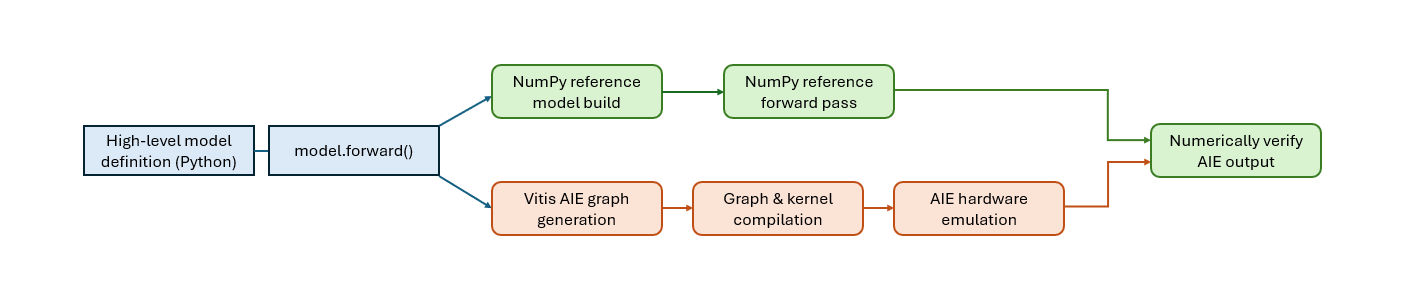}
  \caption{Code-generation and verification flow from \texttt{model.forward()} using the AIEModel framework.}
  \label{fig:codegen-flow}
\end{figure*}

%% file: short-version/background.tex
\section{Prior and Related Work}

Transformer-based architectures such as the Particle Transformer (ParT) have set the state of the art for jet tagging performance in offline analyses~\cite{qu2022parttransformer}. Recent work has begun to explore deploying transformers in real-time trigger environments, including sub-microsecond FPGA implementations for jet tagging using aggressive quantization and specialized dataflows~\cite{laatu2025submicrosecond}. In parallel, several frameworks target transformers or AI workloads on Versal-class architectures and AI Engines, such as the CAT customized transformer accelerator framework~\cite{zhang2024cat}, and Vitis ONNX Execution Provider backends that map high-level models to the AIE array of Ryzen NPUs~\cite{VitisAIExecProvider}. In contrast, CAT jointly exploits both PL and AIE resources, and the ONNX backends focus on Ryzen devices rather than Versal SoCs, whereas our work maps high-level models using an integer-only code-generation framework targeting only the AIE on the VCK190.

%% file: short-version/method1.tex
\section{System Overview}

Our work focuses on building a lightweight code generation framework that maps user-defined model definitions in Python to the AIE of the Versal SoC and developing support within this framework for the quantized computational layers required for the transformer for jet tagging model.

\subsection{Target Platform}

Our implementation targets the AMD Versal VCK190 SoC, with computation mapped exclusively onto the on-chip AI Engine (AIE) array. The AIE tiles provide a spatially distributed, VLIW-style SIMD compute network with local memories and high-bandwidth streaming interconnects, which we exploit to implement the transformer layers. The design is developed using the AMD/Xilinx Vitis 2024.1. 

All neural network computations are performed using int8 quantized, integer-only arithmetic, with int32 accumulators and fixed-point rescaling. At the low level, we implement AIE kernels in C/C++ and connect them using the Vitis dataflow graph programming model, where kernels are nodes and streaming channels between AI Engine tiles are edges. We primarily validate the design using AIE hardware emulation, which allows cycle-accurate evaluation of the generated graph. 

\begin{comment}
\aba{if u have space, put a subsection summarizing the constraints of AIE. like local mem width, bandwiths...etc. Later, refer to that subsection to say how u implmented different things within those constraints}
\end{comment}

\subsection{Code-Generation Framework}
%\zhenghua{This sub-section mostly looks good to me}
Porting nontrivial models onto the AI Engine is
labor-intensive. Developers must hand-write both the C/C++ kernel code and the Vitis graph description that instantiates kernels, wires streams, and places computation across tiles. As the dataflow grows to include multiple attention heads, stacked attention blocks, and residual connections, manually managing ports, FIFOs, and tile workloads becomes error-prone, and debugging mismatches between intended and actual dataflow is difficult. These challenges motivate a code-generation framework that separates high-level model specification from low-level AIE graph construction.

Although a transformer block and the full model appear complex, they are composed of a small set of modular, highly repetitive building blocks: dense projections, multi-head attention, residual additions, and simple nonlinearities. This regular structure is well suited to a templated code-generation approach, where each logical layer encapsulates both its numerical behavior and its hardware mapping.

Our framework exposes a user-facing Python API centered on an \texttt{AIEModel} class, in which a model is defined as a sequence of \texttt{AIELayer} instances (e.g., \texttt{Dense}, \texttt{ResAdd}, \texttt{MHA}), in a style similar to common deep learning libraries. At build time, each \texttt{AIELayer} consumes a set of parameters (tensor shapes, quantization scales, and tiling choices) and emits the corresponding fragments of the Vitis graph, kernel instances, input/output ports, and stream connections. These fragments are then composed into a complete AIE graph, automatically bound to a library of pre-written C/C++ kernels that implement quantized dense layers, residual additions, and MHA. To ensure numerical correctness and simplify debugging, the same high-level Python model definition is also used to construct a NumPy-based reference implementation of the transformer. For any given input, the framework runs both the NumPy ``golden'' model and the AIE implementation and compares the numerical output of the AIE against the reference NumPy implementation.

The core user-facing entry point is the \texttt{model.forward(input)} call. When invoked, this routine (i) constructs the corresponding Vitis AIE graph for the given high-level model, (ii) triggers compilation and AIE hardware emulation to obtain the fixed-point accelerator output, and (iii) in parallel builds an equivalent NumPy reference model and runs a forward pass on the same input. After emulation completes, the framework compares the AIE output against the NumPy reference to validate numerical correctness of the AIE design. Figure~\ref{fig:codegen-flow} sketches this flow, from high-level Python model description through code generation, compilation, AIE emulation, and NumPy-based numerical validation.

The framework is deliberately simple and modular, so that individual layers can be replaced with more optimized kernels, alternative tiling strategies, or richer quantization schemes without changing user-facing model code. Distinct, intermediate model definitions can easily be configured for testing and benchmarking. 

\begin{comment}
We could also have a figure illustrating our AIELayer and AIEModel classes (what they emit, what they compute, object attributes and methods), but I don't think there is space in this report.

Here was the snippet I had before introducing this figure: " A schematic of the software structure illustrates how layer classes, the \texttt{AIEModel}, and the code-generation backend interact to produce the final Vitis graph and kernel code (Fig.~X)."

\end{comment}

%% file: short-version/method2.tex
\section{Architecture and AIE Mapping}
    
\subsection{Jet Tagging Transformer Model}

\begin{figure}
    \centering
    \includegraphics[width=0.45\linewidth]{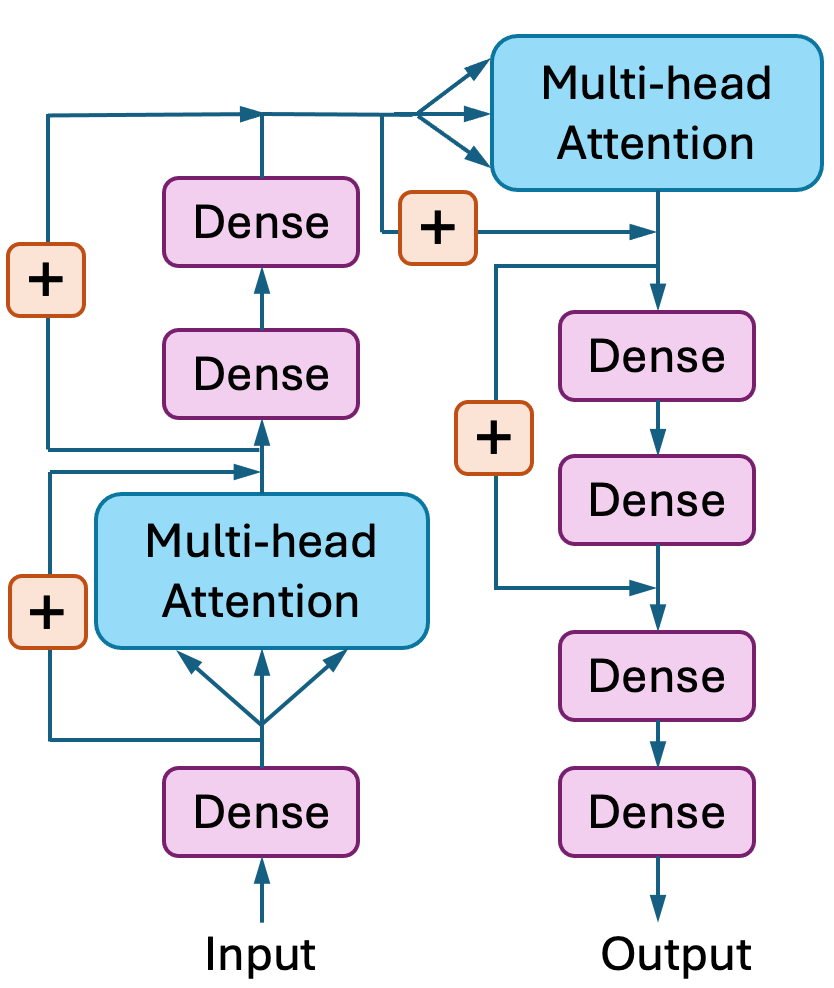}
    \caption{High-level jet tagging transformer model.}
    \label{fig:model-diagram}
\end{figure}

The features and dimensions of the transformer for jet tagging model motivated the development of our framework. The input of the model is an int8 tensor of shape \((160, 8)\) which represents a quantized particle jet after padding for tiling compatibility. Fig.~\ref{fig:model-diagram} shows the design of the network: a compact transformer encoder with two stacked 4-head self-attention (MHA) layers using 57 AIE tiles. Each transformer block consists of an MHA layer followed by a position-wise feed-forward subnetwork with a 64-dimensional hidden layer, implemented using dense layers with ReLU activations. Across the entire model, there are seven dense layers and four residual-add connections, corresponding to the attention and feed-forward sublayers and the final classification head, respectively. We do not apply any normalization layers (e.g., layer normalization or batch normalization). Instead, stability is maintained via the quantization scheme described in Section~\ref{sec:quantization}. Normalization scales can be folded into upstream transforms prior to
quantization~\cite{Jacob2018BNFold}. 
\begin{comment}
Fully integer-only approximations of layer normalization are possible in transformer models, but support for such techniques are left for future work\cite{Kim2021IBERT}.\zhenghua{You can move this to future work. }
\end{comment}

% \begin{figure}[t]
%   \centering
%   \includegraphics[
%     width=\columnwidth,
%     trim=70mm 0mm 70mm 0mm,
%     clip
%   ]{images/model.png}
%   \caption{High-level jet tagging transformer model. }
%   \label{fig:model-diagram}
% \end{figure}

\subsection{AIE Design and Mapping}

\begin{figure}[t]
    \centering
    \includegraphics[width=0.99\linewidth]{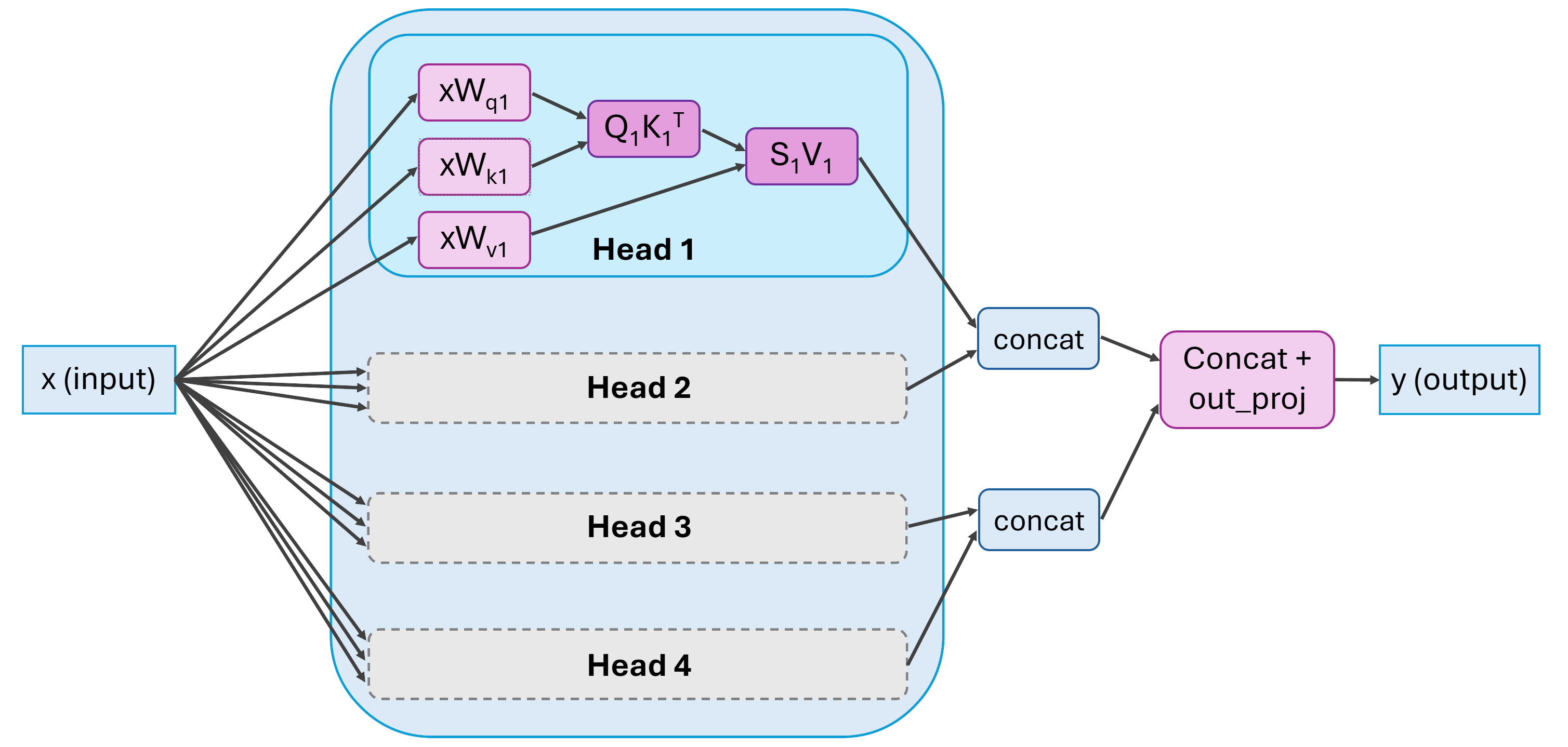}
    \caption{Code-Generated Vitis AIE graph for an MHA layer.  Purple boxes indicate individual compute tiles, and arrows indicate single-stream int8 data connections.}
    \label{fig:mha-design}
\end{figure}

We focus on demonstrating a functional mapping of the transformer to the VCK190 AIE array and on exercising the full end-to-end flow of our code-generation framework. All compute-intensive operations (dense layers, multi-head attention (MHA), and residual additions) are implemented as a set of kernels and mapped to separate AIE tiles.
\begin{comment}
\zhenghua{I swapped these 2 paragraphs. In general, put your important stuff in the from unless they are dependent on sth. Maybe in 1-2 sentences, mention the underlying inefficiencies / complexes in MHA layers here.}
\end{comment}
We introduce an AIE mapping of MHA layers via head-level parallelism. For each attention block, the per-head query, key, and value projections are assigned to separate tiles, so that different head sub-projections can be processed in parallel. After each head computes its contribution, a small set of concatenation kernels collect the per-head outputs and form the full d\_model-dimensional representation, which is then passed through an output projection kernel. The mapping of MHA kernels is illustrated in Figure \ref{fig:mha-design}. This highly parallel design exploits the large tile count of the VCK190 AIE and results in a FIFO latency speedup proportional to the number of heads defined in the MHA layer, as shown in Table~\ref{tab:part_skeleton}.

Dense layers are implemented as single AIE kernels that perform int8 matrix--matrix multiplications with int32 accumulation and bias addition. After accumulation, requantized values are streamed to downstream kernels. Residual connections are realized as separate AIE kernels that consume two int8 input streams and perform elementwise additions.

\subsection{Quantization and Integer-Only Computations}
\label{sec:quantization}

Deploying attention models efficiently on our AIE requires integer operations. This makes both quantization of linear layers and the realization of non-linear operations such as softmax difficult to implement efficiently, since we must control quantization error while conforming to the hardware’s integer-friendly execution model.

All neural network computations on the AIE use symmetric int8 quantization with int32 accumulation. For each weights tensor, we define a per-tensor scale based on the maximum absolute value observed over calibration and map real values to the integer range \([-127, 127]\). To match the hardware datapath, the effective real-valued requantization factor is implemented as a dyadic rational \(M / 2^{b}\), where \(M\) and \(b\) are integers precomputed offline. 

At runtime, the forward pass applies an integer matrix multiplication in int32, followed by a requantization step that multiplies by \(M\) and right-shifts by \(b\) using banker’s rounding (conv\_even). This design keeps all intermediate values in integer formats while closely approximating standard floating-point inference, thereby improving efficiency on the AIE.

Softmax is particularly challenging in this setting because exponential operations are typically implemented in floating point. 
% \aba{if u have space, put the algorithm somewhere}
We adopt an integer-only variant of the Int-Softmax algorithm introduced in I-BERT~\cite{Kim2021IBERT}, which reformulates the exponential into a sequence of fixed-point operations that efficiently map to our integer hardware. For numerical stability, we first compute the maximum and subtract it from each element. We then decompose each shifted input as
\[
x_i = -k_i \ln 2 + r_i
\]
where \(k_i\) is an integer quotient and \(r_i \in [-\ln 2, 0]\). This allows the exponential to be approximated as
\[
e^{x_i}   \approx 2^{-k_i} \cdot p(r_i),
\]
where \(2^{-k_i}\) is implemented as a bit-shift and \(p(r_i)\) is a second-order polynomial. All quantities (\(k_i\), \(r_i\), polynomial coefficients, and intermediate products) are represented in fixed-point integer formats with shared scaling factors. The resulting approximation introduces a maximum error of \(1.3 \times 10^{-3}\), which is small relative to the overall quantization error of the network.

Our quantization and softmax design demonstrates that a fully integer-oriented attention stack is feasible on the AIE. However, our integer softmax introduces an order-of-magnitude increase in latency. Optimizing these implementations for higher throughput is a primary direction for future work.

%% file: short-version/results.tex
\section{Evaluation and Results}

All AIE experiments in this work are conducted with randomly initialized weights and randomly generated inputs, and are intended as implementation validation studies. Quantization is exercised in two modes across the test suite: a dynamic mode, where quantization factors are obtained by calibration from the reference forward pass used primarily for models with randomly initialized weights and inputs, and a static mode, where fixed factors are chosen by the user ahead of time to provide a numerically informative operating range for validation. Under both modes, verification is based on agreement between the AIE implementation and the corresponding integer reference computations, as described in Sec.~\ref{sec:quantization} and illustrated by the code-generation and validation flow in Fig.~\ref{fig:codegen-flow}.

% , so the reported results demonstrate arithmetic and kernel-level correctness, not end-to-end accuracy of a fully trained and deployment-calibrated model.

We first evaluate a simplified ``skeleton'' version of the jet tagging transformer. This model retains the full MHA and feed-forward structure but omits softmax and bias addition, allowing us to isolate the impact of dense projections and MHA on latency and scaling behavior. In this setting, we compare single-head and four-head configurations under dynamic quantization. As shown in Table~\ref{tab:part_skeleton}, head-parallel execution on the AIE yields roughly a $4\times$ reduction in end-to-end graph latency (from \(7.35\times 10^{5}\,\text{ns}\) for 1 head to \(1.87\times 10^{5}\,\text{ns}\) for 4 heads) and a $4\times$ increase in throughput.

\begin{table}[t]
  \centering
  \caption{Latency and throughput for the jet tagging skeleton model
  (no bias, no softmax).}
  \label{tab:part_skeleton}
  \begin{tabular}{lccc}
    \hline
    Config. & Latency (ns) & Throughput (MB/s) & Throughput (samples/s) \\
    \hline
    4 heads & 187014.2  & 12.85 & 10041.8 \\
    1 head  & 734559.2  & 3.55 & 2775.3 \\
    \hline
  \end{tabular}
\end{table}

To understand the performance impact of our integer-only softmax implementation, we conduct a set of benchmarks on a small base model. The base model consists of a single dense layer, which is computationally equivalent to a $160 \times 64 \times 64$ matrix multiplication. We then incrementally augment this model with bias addition and integer-only softmax to measure their effects on runtime latency and throughput. As shown in Table~\ref{tab:softmax_results}, adding bias alone increases latency by nearly an order of magnitude relative to the dense-only baseline, while introducing softmax causes two orders of magnitude slowdown and a corresponding drop in throughput. These results highlight that the integer-only softmax and associated data movement currently form a dominant bottleneck in our design and motivate future work on more efficient kernel implementations and dataflows for these operations within our framework.

\begin{table}[t]
  \centering
  \caption{Latency and throughput for a base test model
  consisting of a single dense layer.}
  \label{tab:softmax_results}
  \begin{tabular}{lcc}
    \hline
    Configuration & Latency (ns) & Throughput (MB/s) \\
    \hline
    Dense (no bias, no softmax)      & 218.3   & 697.663 \\
    Dense + bias (no softmax)        & 1695.8  & 22.408  \\
    Dense + softmax (no bias)        & 55199.2 & 4.603   \\
    Dense + bias + softmax           & 55392.5 & 4.585   \\
    \hline
  \end{tabular}
\end{table}

%% file: short-version/conclusion.tex
\section{Conclusion and Future Work}
In this work, we presented a preliminary implementation of a fully quantized, integer-oriented transformer for jet tagging on the AMD Versal VCK190 AI Engine. This includes a novel, head-parallel mapping of MHA layers onto the AIE grid. Our primary contribution is a lightweight, reusable code-generation framework that maps high-level Python model descriptions onto the AIE as composable building blocks, automatically producing Vitis AIE graphs. 
\begin{comment}
At the same time, our design is deliberately simple and should be viewed as a starting 
point rather than a production-ready accelerator. 
\aba{dont be apologetic}\zhenghua{Be forward-looking and optimistic}
\end{comment}

Key directions for future work include: (i) developing more optimized kernels and graph dataflows to reduce end-to-end latency and mitigate the current softmax and bias bottlenecks; (ii) extending the framework to support additional transformer components, such as integer-only normalization and pooling layers; and (iii) improving the quantization scheme beyond symmetric per-tensor scaling and rigorously evaluating the system with trained jet tagging models and realistic calibration data. The framework’s modular structure is intended to make each of these enhancements incremental and composable, enabling sustained co-design of models, quantization, and AIE mappings.

\begin{comment}
    This section is short (perhaps too short) in an effort to keep everything to 4 pages. If we shorten another section we may be able to make our conclusion longer.
\end{comment}